\documentclass[journal]{IEEEtran}
\ifCLASSINFOpdf
\else
\fi
\usepackage{amsmath}
\usepackage{amsfonts}
\usepackage{amssymb}

\usepackage{graphicx}
\usepackage{epstopdf}

\graphicspath{{./fig/}}
\usepackage{color}

\usepackage{color}
\usepackage{amsmath}
\usepackage{algorithm, algorithmic,subfig,amsfonts,makecell,array}
\newcommand{\myPara}[1]{\vspace{.05in}\noindent\textbf{#1}}

\newcommand{\mr}[1]{\mathrm{#1}}

\usepackage{float}

\def\eg{{\em e.g.}}
\def\etal{{\em et al.}}

\newcommand{\bm}[1]{\mbox{\boldmath{$#1$}}}

{}
{}
{}

\begin{document}
%
\title{Student Network Learning via Evolutionary Knowledge Distillation}
%
%
%

\author{Kangkai Zhang, 
        Chunhui Zhang,
        Shikun Li,
        Dan Zeng,~\IEEEmembership{Member,~IEEE,}
        and Shiming~Ge,~\IEEEmembership{Senior Member,~IEEE}


\thanks{K.~Zhang, C.~Zhang, S.~Li and S.~Ge are with the Institute of Information Engineering, Chinese
Academy of Sciences, Beijing 100095, China, and School of Cyber Security, University of Chinese
Academy of Sciences, Beijing 100049, China. Email: \{zhangkangkai, zhangchunhui, lishikun, geshiming\}@iie.ac.cn.}
\thanks{D.~Zeng is with the Department of Communication Engineering, Shanghai University, Shanghai 200444, China. E-mail: dzeng@shu.edu.cn.}

\thanks{Shiming Ge is the corresponding author. E-mail: geshiming@iie.ac.cn.}}

\maketitle

\begin{abstract}
Knowledge distillation provides an effective way to transfer knowledge via teacher-student learning, where most existing distillation approaches apply a fixed pre-trained model as teacher to supervise the learning of student network. This manner usually brings in a big capability gap between teacher and student networks during learning. Recent researches have observed that a small teacher-student capability gap can facilitate knowledge transfer. Inspired by that, we propose an evolutionary knowledge distillation approach to improve the transfer effectiveness of teacher knowledge. Instead of a fixed pre-trained teacher, an evolutionary teacher is learned online and consistently transfers intermediate knowledge to supervise student network learning on-the-fly. To enhance intermediate knowledge representation and mimicking, several simple guided modules are introduced between corresponding teacher-student blocks. In this way, the student can simultaneously obtain rich internal knowledge and capture its growth process, leading to effective student network learning. Extensive experiments clearly demonstrate the effectiveness of our approach as well as good adaptability in the low-resolution and few-sample visual recognition scenarios.
\end{abstract}


%
\IEEEpeerreviewmaketitle

\section{Introduction}
\IEEEPARstart{D}{eep} neural networks have proved success in many visual tasks like image classification~\cite{Matiur18, Takahash20}, object detection~\cite{chen2017learning, chen2017pixelwise}, semantic segmentation~\cite{ghosh2019understanding} and other fileds~\cite{zhao2019scale, hu2019channel, zhai21} due to their powerful knowledge extraction capability from massive available data. Beyond these remarkable successes, there is still increasing concern that the development of effective learning approaches for the real-world scenarios where the high-quality data often is not available or insufficient. In this case, network learning will encounter obstacles. Knowledge distillation \cite{Hinton2015DistillingTK} provides an economic way that can transfer knowledge from a pre-trained teacher network to facilitate the learning of a new student network.

\begin{figure}[t]
	\centering
	\includegraphics[width=1.0\linewidth]{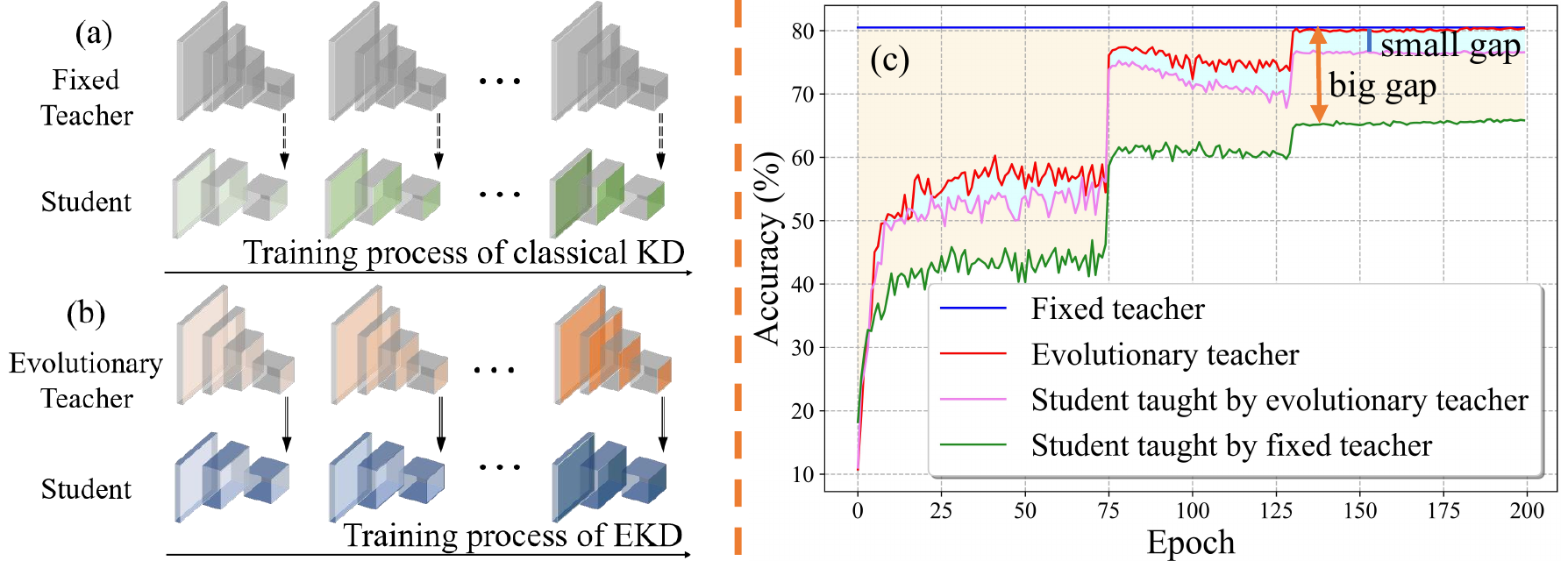}
	\caption{Unlike classical knowledge distillation (KD) approaches with fixed teacher, an evolutionary teacher can enable more efficient knowledge transfer by minimizing the capability gap between teacher and student. Thus, we propose evolutionary knowledge distillation (EKD) to facilitate student network learning.}
	\label{fig:fig1_motivation}
\end{figure}

Knowledge distillation approaches mainly follow offline or online strategies. Offline strategies try to design more effective knowledge representation methods to learn from powerful teacher. Park \etal~\cite{park2019relational} and Liu \etal~\cite{liu2019knowledge} focused on the structured knowledge about the instances, while other approaches~\cite{RomeroBKCGB14, tian2019crd, yim2017gift} paid attention to some internal information in the network during distillation process. These offline approaches usually use a fixed pre-trained model as teacher, as shown in Fig.~\ref{fig:fig1_motivation} (a), and this manner usually brings in a big capability gap between teacher and student during learning (see Fig.~\ref{fig:fig1_motivation} (c)), leading to transfer difficulties. The capability gap refers to the performance difference between teacher and student network, in the image classification task, it specifically refers to the difference in accuracy. By contrast, the online distillation strategies attempt to reduce the capability gap by some training schemes for the absence of pre-trained teachers to improve the learning of student. An on-the-fly native ensemble (ONE) scheme~\cite{LanZG18} was proposed for one-stage online distillation. Instead of borrowing supervision signals from previous generations, snapshot distillation~\cite{YangXSY19} extracted information from earlier iteration in the same generation. These online practices provide some good solutions to the shackles of capability gap brought by the fixed teacher, resulting in relatively better knowledge transfer, while they need high demands on the efficiency of knowledge representation because they rely more on their relatively less reliable peers' or their own predictions to provide additional supervision~\cite{LanZG18, AnilPPODH18, chen2020online}. Hence, there is a question: \emph{how to ensure both small capability gap and efficient knowledge representation to facilitate student learning?}

Inspired by the recent observations~\cite{YangXSY19, mirzadeh2019improved, jin2019knowledge} that a small teacher-student capability gap is beneficial to knowledge transfer, we propose an evolutionary knowledge distillation (EKD) approach to improve the learning of student, as shown in Fig.~\ref{fig:fig1_motivation} (b). The approach uses an evolutionary teacher whose performance is continuously improved as the training process to constantly transfer intermediate knowledge to the student, the evolutionary teacher could provide richer supervision information for the learning of student and reduce the capability gap. In addition, to improve the knowledge representation ability, the teacher and student networks are both divided into several blocks, and a simple guided module pair is introduced between each corresponding block. 
In short, the evolutionary teacher not only solves the problem of capability gap between fixed teacher and student of offline knowledge distillation, but also solves the problem of insufficient and relatively unreliable supervision information caused by the absence of a qualified teacher for online knowledge distillation. In addition, the guided modules further promote the representation of intermediate knowledge. In this way, the student can continuously and adequately learn the intermediate knowledge from teacher as well as its growth process. 

The main contributions of our work could be summarized in three folds: 
1)~We propose an evolutionary knowledge distillation approach to facilitate the student network learning by narrowing the teacher-student capability gap;
2)~We introduce guided module pairs to enhance the knowledge representation and transfer ability;
3)~We conduct extensive experiments to verify that our approach is superior to the state-of-the-arts and exhibits better adaptability in the low-resolution and few-sample scenarios.

\begin{figure*}[t]
	\centering
	\includegraphics[width=0.95\linewidth]{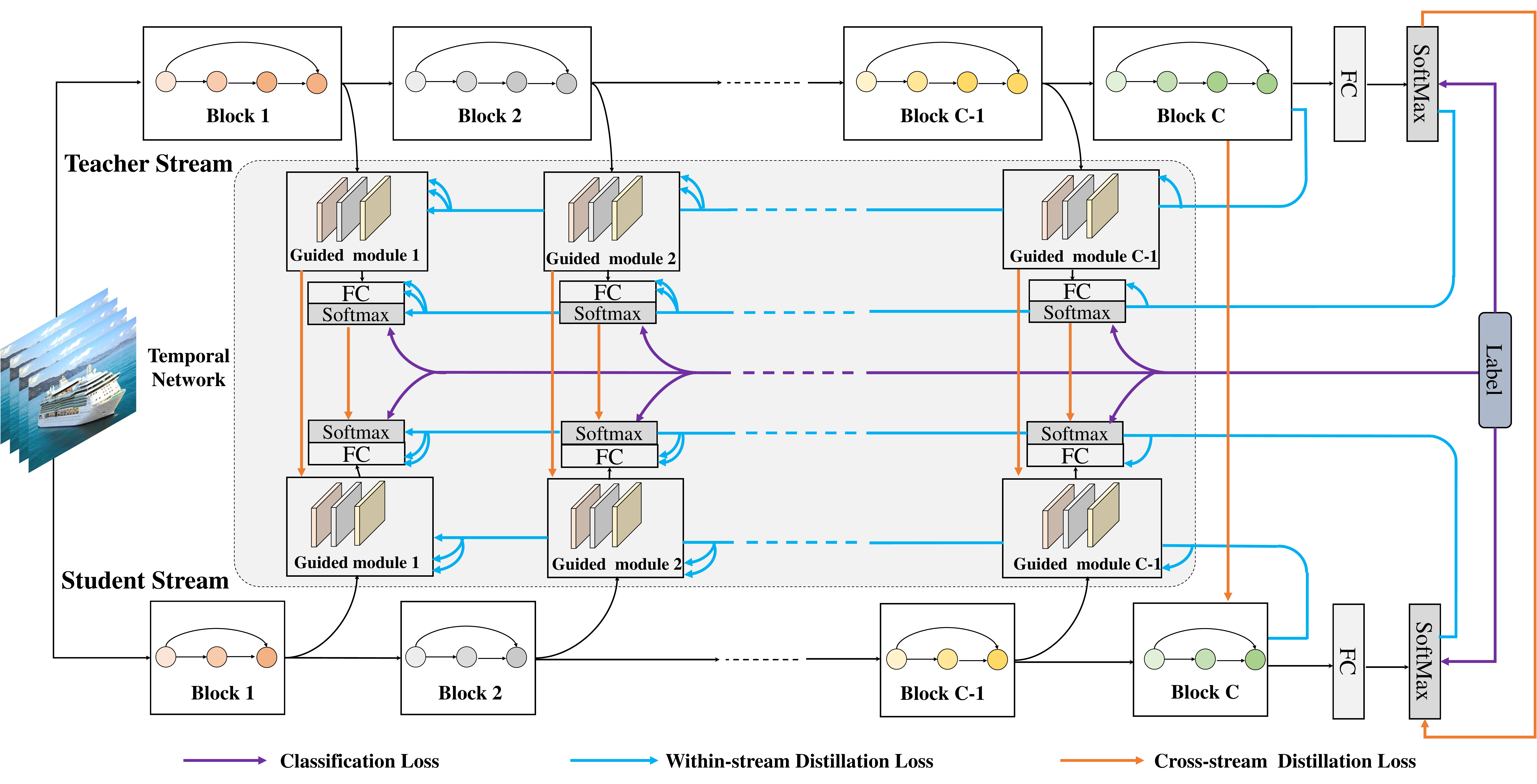}
	\caption{The network details of each batch data process of our evolutionary knowledge distillation approach. We divide the teacher stream and student stream into $ C $ blocks, between each block of teacher and student, a pair of guided modules is introduced to help knowledge representation and transfer, so that the student can effectively capture both \textbf{what} knowledge the teacher learns and \textbf{how} it grows.} 
	\label{fig2:framework}
\end{figure*}

\section{Related Work}
\subsection{The Basic of Knowledge Distillation}
Knowledge distillation (KD) provides a concise but effective solution for transferring knowledge from a pre-trained large teacher model to a smaller student model~\cite{Hinton2015DistillingTK}. Since the introduction of knowledge distillation, it has been widely used in image recognition, semantic segmentation, and other fields, especially model compression~\cite{Hinton2015DistillingTK, bucilu2006model, BaC14, PolinoPA18, LiuCLQLW19, liu2021deep}. In practice, the student model learns the prediction of pre-trained teacher model to make itself more powerful than it's trained alone. Compared to hard ground-truth labels, fine-grained class information in soft predictions helps the small student model to reach flatter local minima, which results in more robust performance and improves generalization ability~\cite{PereyraTCKH17, KeskarMNST17}. Several recent works attempt to further improve that transfer knowledge between varying-capacity network models with offline or online knowledge distillation approaches~\cite{tian2019crd, LanZG18, chen2020online, ying2018DML, AhnHDLD19, ChenZD18}.

\subsection{Offline Knowledge Distillation}
The offline knowledge distillation approaches often adopt a two-stage training mode, it first trains the teacher model, and then trains the student network by various distillation strategies. Classical FitNet~\cite{RomeroBKCGB14} tried to transfer more supervision information by using the feature map of the teacher network middle layer firstly. Crowley \etal~\cite{CrowleyGS18} proposed structural model distillation for memory reduction using a strategy that produced a student architecture that was a simple transformation of the teacher's: no redesign is needed, and the same hyper-parameters can be used. And some recent approaches~\cite{liu2019knowledge, 9157683} attempted to pay more attention to the relationship information of the instances. Tian \etal~\cite{tian2019crd} proposed contrastive representation distillation, the main idea is very general: learn a representation that is close in some metric space for ``positive'' pairs and push apart the representation between ``negative'' pairs. Furlanello~\etal~\cite{furlanello2018born} interactively absorbed the distilled student models into the teacher model group, through which the better generalization ability on test data is obtained. Sukmin Yun \etal~\cite{yun2020regularizing} proposed a new regularization method that penalizes the predictive distribution between similar samples via self knowledge distillation to mitigate the issue that deep neural networks with millions of parameters may suffer from poor generalization due to overfitting.~\cite{you2017learning, ruder2017knowledge, shen2019meal, shen2020meal} proposed utilizing multiple teachers to provide more supervision for the learning of student network.

Generally speaking, for offline knowledge distillation, in order to achieve a great effect, the following conditions need to be met: a high-quality pre-trained teacher model, difference between teacher and student, adequate supervision information~\cite{yim2017gift, YangXSY19, yang2018knowledge}. In particular, offline knowledge  distillation methods rely heavily on a fixed pre-trained model, however, the huge capability gap between fixed teacher and student model will bring great challenges to knowledge transfer. To bridge this gap, Mirzadeh \etal~\cite{mirzadeh2019improved} introduced multi-step knowledge distillation, which employed an intermediate-sized network (teacher assistant); Jin \etal~proposed a method named RCO~\cite{jin2019knowledge}, which utilizes the route in parameter space teacher network passed by as a constraint to bring a better optimization to student. None of these methods can completely solve the obstacles of knowledge transfer caused by the capability gap due to the fact that they also rely to some extent on the pre-trained teacher model. Therefore, we need to rethink how to break the shackles of offline distillation approaches to improve the learning of student network more effectively.

\subsection{Online Knowledge Distillation}
Different from the conventional two-stage offline distillation approaches, the current approaches increasingly focus on the online distillation strategies, which attempt to reduce the capability gap by some training schemes in the absence of pre-trained teachers to improve the learning of student. A group of networks or sub-networks will be trained almost synchronously, which aims to improve the performance of student by using the predictions of their peers or sub-networks as supervision instead of the high-quality pre-trained teachers'. Deep Mutual Learning (DML)~\cite{ying2018DML} applied distillation losses mutually, treating each other as teachers, and it achieved good results. However, DML lacks an appropriate teacher role, hence provides only limited information to each network. Guocong Song \etal~\cite{SongC18} introduced collaborative learning in which multiple classifier heads of the same network are simultaneously trained on the same training data to improve generalization and robustness to label noise with no extra inference cost. A similar learning strategy named On-the-fly Native Ensemble (ONE) for one-stage online distillation proposed by Xu Lan~\etal~\cite{LanZG18}. Specifically, ONE trains only a single multi-branch network while simultaneously establishing a strong teacher on-the-fly to enhance the learning of target network. Chung~\etal~\cite{chung2020feature} proposed an online knowledge distillation method that transfers the knowledge of the class probabilities and the feature map using the adversarial training framework. Zhang \etal~\cite{ZhangSGCBM19} proposed an online training framework called self-distillation, which forces student to refine its knowledge inside the network, thereby improving itself. And a framework Snapshot Distillation was proposed for teacher-student optimization in one generation~\cite{YangXSY19}, which extracted such information from earlier epochs in the same generation, meanwhile made sure that the difference between teacher and student is sufficiently large so as to prevent under-fitting. After these, a novel two-level framework OKDDip~\cite{chen2020online} was proposed to perform distillation during training with multiple auxiliary peers and one group leader for effective online distillation. In OKDDip framework, the first-level distillation works as diversity maintained group distillation with several auxiliary peers, while the second-level distillation transfers the diversity enhanced group knowledge to the ultimate student model called group leader.

These online, timely and efficient training methods are promising and some good progress has been made in narrowing the capability gap between teacher and student model~\cite{LanZG18, YangXSY19, ZhangSGCBM19}. However, for online distillation, two factors still hold them back. First, although the teachers of online distillation methods are dynamic and have narrowed the capability gap, the gap still exists due to the lack of representation in detail and the process of learning. Second, owing to the absence of a qualified teacher role for online knowledge distillation, the insufficient and relatively unreliable supervision information will restrict the learning of student to some extent. There is still great room for improvement in knowledge transfer and representation. Therefore, we propose the evolutionary knowledge distillation approach to enhance the performance of student network learning by using an evolutionary teacher and focusing on intermediate knowledge of the teaching process dynamically.

\section{The Proposed Approach}\label{sec3}
In our evolutionary knowledge distillation (EKD) approach, the teacher and student network are trained almost synchronously, and an evolutionary teacher can provide supervision information for the learning of student. The performance of teacher is continuously improved as the training process, which can provide richer supervision information and reduce the capability gap. As shown in Fig.~\ref{fig2:framework}, EKD consists of teacher stream, student stream and some extra guided modules. The network of each stream is divided into several blocks, depending on the specific network. Previous experiences have shown that using early blocks helps to exploit information inside the network~\cite{AnilPPODH18, ying2018DML, HeZRS16}, in order to utilize the intermediate knowledge, we improve bottleneck module~\cite{HeZRS16, Cheng20} to form some guided module pairs that are applied to assist the representation and transfer of knowledge. For each corresponding block between teacher and student stream, a guided module pair is inserted to make distillation process effective and efficient within stream and cross stream. Within-stream knowledge distillation aims to enhance knowledge representation, while cross-stream knowledge distillation helps to improve knowledge transfer.

\begin{algorithm}
	\renewcommand{\algorithmicrequire}{\textbf{Input:}}
	\renewcommand{\algorithmicensure}{\textbf{Output:}}
	\caption{Student network learning via EKD}
	\label{alg:alg1}
	\begin{algorithmic}[1]
		\REQUIRE Training set $\mathbb{D}=\{(\bm{x}_i, \bm{y}_i)\}_{i=1}^{|\mathbb{D}|}$;
		\ENSURE Optimized student network weight: $ \bm{\mr{w}}_s$;
		\STATE Initialize the Teacher and Student;
		\FOR{The student model did not converge}
		\STATE Get a batch of data;
		\STATE Feed the data into \textbf{teacher} stream;
		\STATE Get the feature map, soften probabilities of the teacher stream;
		\STATE Compute the with-stream distillation loss with Eq.~(\ref{within loss}) and Eq.~(\ref{Feature loss});
		\STATE Compute the classification loss with Eq.~(\ref{classification loss});
		\STATE Compute the total loss of teacher with Eq.~(\ref{total});
		\STATE Compute gradient to model parameters $\bm{\mr{w}_t}$ and update with the SGD optimizer.
		\STATE Feed the data into \textbf{student} stream;
		\STATE Get the feature map, soften probabilities of the student stream;
		\STATE Compute the with-stream distillation loss with Eq.~(\ref{within loss}) and Eq.~(\ref{Feature loss});
		\STATE Compute the cross-stream distillation loss with Eq.~(\ref{Guided1}) and Eq.~(\ref{Guided2});
		\STATE Compute the classification loss with Eq.~(\ref{classification loss});
		\STATE Compute the total loss of student with Eq.~(\ref{total});
		\STATE Compute gradient to model parameters $\bm{\mr{w}_s}$ and update with the SGD optimizer.
		\ENDFOR
		\STATE \textbf{Return} the optimized student network weight $\bm{\mr{w}_s}$.
	\end{algorithmic}
\end{algorithm}

\subsection{Problem Formulation}
Denote the training set is $\mathbb{D}=\{(\bm{x}_i, \bm{y}_i)\}_{i=1}^{|\mathbb{D}|}$, where $\bm{x}_i$ is the $i$th sample with label $\bm{y}_i\in\{1,2,...,m\}$. Here $m$ is the number of classes. The objective of distillation is transferring the knowledge of teacher $\phi_t(\bm{x};\bm{\mr{w}}_t)$ into the student $\phi_s(\bm{x};\bm{\mr{w}}_s)$, where $\bm{\mr{w}}_t$ and $\bm{\mr{w}}_s$ are the model parameters of teacher and student, respectively. In our setting, the network of each stream is divided into $C$ blocks. Between the corresponding teacher and student blocks, a pair of guided modules $\mathcal{M}_b=\left(\phi_g(\bm{f}^{(b)}_{t};\bm{\mr{w}}^{(b)}_{t}),\phi_g(\bm{f}^{(b)}_{s};\bm{\mr{w}}^{(b)}_{s})\right)$ is introduced to assist the representation, transfer of knowledge, where $\bm{f}^{(b)}_{k}$ and $\bm{\mr{w}}^{(b)}_{k}$ are the feature maps and model parameters from the $b$th $(b=1,...,C-1)$ block of teacher ($k=t$) and student ($k=s$) respectively. Therefore, during student network learning, the problem of EKD is formulated as:
\begin{equation}\label{}
\setlength{\abovedisplayskip}{3pt}
\begin{aligned}
\min_{\bm{\mr{w}}_t,\bm{\mr{w}}_s,\{\bm{\mr{w}}^{(b)}_{t},\bm{\mr{w}}^{(b)}_{s}\}}\mathcal{L}(\phi_t(\bm{x};\bm{\mr{w}}_t), \{\mathcal{M}_b\}^{C-1}_{b=1},\phi_s(\bm{x};\bm{\mr{w}}_s)),
\end{aligned}
\end{equation}
where $\mathcal{L}$ is the function to measure distillation loss. Different from traditional knowledge distillation where $\phi_t$ is fixed, EKD simultaneously learns both $\phi_t$ and $\phi_s$ from $\mathbb{D}$ with the help of several guided module pairs $\{\mathcal{M}_b\}^{C-1}_{b=1}$. After learning, the temporal network within the dashed box (see the Fig.~\ref{fig2:framework}) will be discarded and only the parameters of the main model of student stream $\bm{\mr{w}_s}$ will be used for inference.

\subsection{Evolutionary Distillation}
The proposed evolutionary knowledge distillation adopts online training of teacher and student model synchronously, the parameters of student and teacher models are updated in each batch data process, which can reduce the teacher-student capability gap. When coupled with Fig.~\ref{fig2:framework}, we can see that teacher and student streams are divided into $C$ blocks. Each block followed by a guided module with a fully connected layer constitutes multiple classifiers. We assume a stream with $C$ classifiers. The training process includes two simultaneous stages, within-stream distillation and cross-stream distillation. For within-stream distillation, deeper classifiers provide supervision to help the learning of shallow classifiers, which can improve the ability of the stream itself to represent knowledge. Cross-stream distillation can improve knowledge transfer from the evolutionary teacher to student. The guided modules can help facilitate knowledge representation and transfer.

\myPara{Within-stream Distillation.} For within-stream distillation, we use the shallower classifiers $\phi_g(\bm{f_k^{(b)}};\bm{\mr{w}}_k^{(b)}), b \in \{1,...,C-1\}$ as students, and use the deeper classifiers $\phi_k(\bm{x};\bm{\mr{w}}_k)$ and $\phi_g(\bm{f_k^{(c)}};\bm{\mr{w}}_k^{(c)})$, here $ c \in \{b+1,...,C-1\} $, as teachers. The students are trained by the supervision provided by teachers, supervision includes Kullback-Leibler (KL) divergence~\cite{Hinton2015DistillingTK} and L2 distance~\cite{RomeroBKCGB14} between the feature maps before the final FC layers of teacher and student. In order to simplify the form, here we only show the loss between the backbone network $\phi_k(\bm{x};\bm{\mr{w}}_k)$ and each guided modules $\phi_g(\bm{f_k^{(b)}};\bm{\mr{w}}_k^{(b)})$.
\begin{equation}\label{}
\min_{\bm{\mr{w}}_k^{(b)},\bm{\mr{w}}_k}~{\mathcal L}_{w} (\phi_g(\bm{f_k^{(b)}};\bm{\mr{w}}_k^{(b)}), \phi_k(\bm{x};\bm{\mr{w}}_k)), b \in \{1,...,C-1\}.
\end{equation}

In practice, we use the classic distillation loss, KL divergence, and L2 distance loss. We compute KL divergence loss between deep classifiers and shallow classifiers within stream. The output of teacher and student is $\phi_k(\bm{x};\bm{\mr{w}}_k)$ and $\phi_g(\bm{f_k^{(b)}};\bm{\mr{w}}_k^{(b)})$ respectively, $b \in \{1,...,C-1\}$. And the distillation loss $\mathcal L_{D}$ is calculated by the following.
\begin{equation}\label{within loss}
\mathcal L_{D}=\sum_{b=1}^{C-1} \ell(\phi_g(\bm{f_k^{(b)}};\bm{\mr{w}}_k^{(b)}), \phi_k(\bm{x};\bm{\mr{w}}_k)),
\end{equation}
where $\ell$ denotes the KL divergence loss. As multiple different teachers provide different knowledge, we could achieve the more robust and accurate knowledge representation.

We compute feature loss $\mathcal L_{F}$ by L2 distance which can be obtained through computing the feature maps of deep classifier and shallow classifiers.
\begin{equation}\label{Feature loss}
\mathcal L_{F}= \sum_{b=1}^{C-1} \left\| F_{k} - F_{g,k}^{(b)}\right\|_{2}^{2},
\end{equation}
where $F_k$ and $F_{g, k}^{(b)}$ represent the feature maps of teacher or student before the FC layer respectively.

\myPara{Cross-stream Distillation.} For the cross stream knowledge transfer, the student will learn under the supervision of the corresponding guided modules of the teacher stream:
\begin{equation}\label{}
\min_{\bm{\mr{w}}_t^{(c)},\bm{\mr{w}}_s^{(c)}}~{\mathcal L}_{b}(\phi_t(\bm{x};\bm{\mr{w}}_t^{(c)}), \phi_s(\bm{x};\bm{\mr{w}}_s^{(c)})), c \in \{1,...,C\}.
\end{equation}

The cross-stream distillation loss contains two types of losses: distillation loss and feature loss. For distillation loss between backbone networks and the one between each pair of guided modules, they were calculated by the KL divergence.
\begin{equation}\label{Guided1}
\begin{aligned}
\mathcal L_{G1}= & \ell (\phi_t(\bm{x};\bm{\mr{w}}_t), \phi_s(\bm{x};\bm{\mr{w}}_s)) \\& + \sum_{b=1}^{C-1}\ell (\phi_g(\bm{f_t^{(b)}};\bm{\mr{w}}_t^{(b)}), \phi_g(\bm{f_s^{(b)}};\bm{\mr{w}}_s^{(b)})),
\end{aligned}
\end{equation}
where the outputs of guided module attached to the teacher and student are $\phi_g(\bm{f_t^{(b)}};\bm{\mr{w}}_t^{(b)})$ and $\phi_g(\bm{f_s^{(b)}};\bm{\mr{w}}_s^{(b)})$, respectively, $ b\in \{1,...,C-1\} $. The $\phi_t$ and $\phi_s$ denote the outputs of the backbone networks.

The form of feature loss is as follows,
\begin{equation}\label{Guided2}
\mathcal L_{G2}=\sum_{b=1}^{C-1}\left\|F_{g,t}^{(b)}-F_{g,s}^{(b)}\right\|_{2}^{2},
\end{equation}
where $F_{g,t}^{(b)}$ and $F_{g,s}^{(b)}$ represent the feature map of classifier before the FC layer respectively. The feature loss $\mathcal L_{F}$ between each pair of guided modules is to measure differences in the feature map, which can promote intermediate knowledge transfer cross stream. However, we divide the stream of teacher and student into several blocks, the output dimensions between each corresponding block may not match. Through the processing of the guided modules, we can ensure the consistency of the dimensions while ensuring the least loss of intermediate knowledge during the transfer process. 

Then, we combine the two parts of the guided loss, 
\begin{equation}\label{Guided}
\mathcal L_{G}= \mathcal L_{G1} + \mathcal L_{G2}.
\end{equation}

For each classifier, we compute cross entropy loss $\ell_{ce}$ between $\phi(\bm{x};\bm{\mr{w}}^{(c)})$ and $\bm{y}$. In this way, the label $\bm{y}$ directs each classifier's probability as possible. There are multiple classifiers and we sum each cross entropy loss as follows:
\begin{equation}\label{classification loss}
\mathcal L_{L}=\sum_{c=1}^{C} \ell_{ce}\left(\phi(\bm{x};\bm{\mr{w}}^{(c)}), \bm{y}\right).
\end{equation}

As the parameters of teacher and student are constantly updated in each iteration, the evolutionary knowledge distillation process can be formulated as:
\begin{equation}\label{}
\begin{aligned}
\min_{\bm{\mr{w}}_t,\bm{\mr{w}}_s,\bm{\mr{w}}_t^{(b)},\bm{\mr{w}}_s^{(b)}}~{\mathcal L^e}&=\sum {\mathcal L}_w^e + \sum {\mathcal L}_b^e,
\end{aligned}
\end{equation}
where $e$ denotes the number of iterations. Specifically, for teacher stream and student stream, we used different loss functions, $\mathcal L^{(t)}$ and $\mathcal L^{(s)}$, respectively.
\begin{equation}\label{total}
\mathcal L^{(k)} =  \mathcal L_{D} +  \mathcal L_{F} +  \mathcal L_{L} + \delta(k=t) \mathcal L_{G},
\end{equation}
where $\delta(k=t)$ is an indicator function which equals 1 if the stream $k$ is teacher and 0 otherwise.
This means that the training of teacher and student streams are similar but different. The student learns from evolutionary teacher constantly with the supervision provided by teacher network and guided modules. For each iteration, we first optimize a teacher network to provide supervision to the learning of student, after that, the optimization of teacher and student will be carried out synchronously.
In general, the evolutionary teacher reduces the capability gap between teacher and student. The guided modules not only facilitate the knowledge representation of teacher and student stream but also promote the transfer of more intermediate and process knowledge from teacher to student.

\begin{table*}[t]
	\centering
	\caption{Classification accuracy with peer-architecture setting on CIFAR100.}
	\label{tab:similararch}
	\small
	\renewcommand{\arraystretch}{1.1}
	\begin{tabular}{l|ccccccc}
		\hline
		Teacher Network &VGG19(bn) &VGG11(bn)  & ResNet50 & ResNet18 & ResNet101 & ResNet50  & WRN50-2 \\
		Student Network &VGG11(bn)  &VGG11(bn) & ResNet18 & ResNet18  & ResNet50 & ResNet50  & WRN50-2 \\
		\hline
		Teacher Acc. (\%) & 74.17 & 70.76  & 78.71  & 76.61 & 80.05 & 78.71 & 80.24   \\
		Student Acc. (\%) & 70.76 & 70.76  & 76.61  & 76.61 & 78.71 & 78.71 & 80.24   \\
		\hline
		KD~\cite{Hinton2015DistillingTK} & 73.28  & 72.12 & 77.57 & 78.97  & 79.16 & 79.54 & 80.41 \\
		Fitnet~\cite{RomeroBKCGB14}  & 71.51  & 71.00 & 76.59 & 77.58 & 79.76 & 79.18  & 80.84 \\
		Attention~\cite{ZagoruykoK17} & 72.94  & 70.09 & 76.82 &   77.37 & 80.01 & 78.81  & 80.58 \\
		Factor~\cite{KimPK18} 				& 68.12  & 68.12 & 75.26 & 73.81 & 77.54 &  73.07 & 80.12 \\
		PKT~\cite{pkt_eccv} 				 & 73.15 &  72.43  & 78.44 & 78.51 & 79.51 & 79.66  & 80.85 \\
		RKD~\cite{park2019relational} & 72.81  & 71.88 & 78.21 & 78.10 & 80.07 & 78.73  & 80.40 \\
		Similarity~\cite{TungM19} 		& 73.49 & 72.10 & 78.71 & 78.55 & 80.39 & 79.44   & 80.87\\
		Correlation~\cite{peng2019correlation} & 70.90  & 71.35 & 77.48 & 77.83 & 78.17 & 77.18  & 80.37 \\
		VID~\cite{ahn2019variational} 				& 72.83 & 72.32 & 77.71 &  78.16 & 78.17 &76.76 & 80.88 \\
		Abound~\cite{heo2019knowledge}  & 71.88 & 71.07 & 77.70 & 77.47 & 79.02 & 78.46 & 80.95 \\
		CRD~\cite{tian2019crd} & 71.73 & 73.00 & 78.92 &  78.18 & 80.25 & 79.45 & 81.14 \\
		\textbf{EKD~(Ours)}&  \textbf{74.12}  & \textbf{74.52} & \textbf{82.05} &  \textbf{80.74}  & \textbf{81.91} & \textbf{82.48} & \textbf{82.53} \\
		\hline
	\end{tabular}
\end{table*}

\begin{table*}[t]
	\centering
	\small
	\caption{Classification accuracy with cross-architecture setting on CIFAR100.}
	\label{tab:crossarch}
	\renewcommand{\arraystretch}{1.1}
	\begin{tabular}{l|ccccc}
		\hline
		Teacher Network & ResNet18 & VGG11(bn) &  ResNet18 &WRN50-2 & WRN50-2\\
		Student Network & VGG8(bn) & ShuffleNetV1 & ShuffleNetV2 & ShuffleNetV1 &  VGG8(bn)\\
		\hline
		Teacher Acc. (\%) &  76.61 & 70.76 &  76.61  &  80.24 &80.24\\
		Student Acc. (\%) & 69.21 & 66.18 & 70.48  &66.18 &  69.21\\
		\hline
		KD~\cite{Hinton2015DistillingTK} & 71.17 & 72.40 & 75.03 & 71.78  & 70.31\\
		Fitnet~\cite{RomeroBKCGB14} & 70.59 & 70.50 &  72.24 & 70.46 &70.04\\
	Attention~\cite{ZagoruykoK17} & 71.62 & 69.64 &  73.83 & 70.55 & 69.78 \\
		Factor~\cite{KimPK18} & 68.06 & 68.16 &  69.99 & 70.51 & 70.12\\
		PKT~\cite{pkt_eccv} & 72.74 & 72.06 &  74.31 & 69.80 &69.76\\
		RKD~\cite{park2019relational}  & 71.03 & 70.92 &  73.26 & 70.58 & 70.41\\
		Similarity~\cite{TungM19} & 73.07 & 72.31 &  74.95 & 70.70 & 70.00\\
		Correlation~\cite{peng2019correlation} & 69.82 & 70.70 &  72.21 & 70.66 & 69.96 \\
		VID~\cite{ahn2019variational} & 71.75 & 70.59 &  72.07 & 71.61 & 71.00 \\
		Abound~\cite{heo2019knowledge} & 70.42 & 72.56 &  74.64 &  \textbf{74.28} & 69.81 \\
		CRD~\cite{tian2019crd} & 73.17 & 72.38 & 74.88 & 71.08 &  72.50 \\
		\textbf{EKD~(Ours)} & \textbf{73.82}& \textbf{73.18} & \textbf{75.26} & 73.61 & \textbf{74.05}\\
		\hline
	\end{tabular}
\end{table*}

\subsection{Implementation Details}
The detailed training of the EKD is shown in Algorithm~\ref{alg:alg1}. We introduce several identical guided modules on each stream, and each module is followed by a fully connected layer and a softmax layer. In training process, the two networks are trained at the same time, and they will adopt a certain degree of randomization in order to ensure they are not completely synchronized. This kind of incomplete synchronization can provide enough second-hand knowledge to promote student's learning~\cite{yang2018knowledge, YangXSY19}. Once the training is completed, we only keep the student's parameters of the backbone network for inference. 

The experimental results of other distillation methods are based on the open source code of CRD~\cite{tian2019crd}, we implement experiments according to their hyper-parameters settings. The networks are trained with SGD optimizer~\cite{bottou2012stochastic}. The hyper-parameters are set as follows, batch size is 64, the number of threads is 8, the initial learning rate is 0.1, and it will be multiplied by 0.1 when the epoch is equal to 75,130 and 180, respectively. We fix the random seed to 5 and the temperature $T$ of distillation~\cite{Hinton2015DistillingTK} to 4. All the experiments are implemented by PyTorch on a NVIDIA TITAN GPU.

\section{Experiments}\label{sec4}
\subsection{Experimental Settings}
To verify the effectiveness and adaptability of our EKD approach, we conduct experiments on five benchmarks, including CIFAR10~\cite{krizhevsky2009learning}, CIFAR100~\cite{krizhevsky2009learning}, Tiny-ImageNet~\cite{le2015tiny}, UMDFaces~\cite{bansal2017umdfaces} and UCCS~\cite{uccs}. We first compare EKD with several state-of-the-arts offline distillation approaches, including KD~\cite{Hinton2015DistillingTK}, Fitnet~\cite{RomeroBKCGB14}, Attention~\cite{ZagoruykoK17}, Factor~\cite{KimPK18}, PKT~\cite{pkt_eccv}, RKD~\cite{park2019relational}, Similarity~\cite{TungM19}, Correlation~\cite{peng2019correlation}, VID~\cite{ahn2019variational}, Abound~\cite{heo2019knowledge}, CRD~\cite{tian2019crd}, Res-KD~\cite{li2020reskd} and AFD~\cite{ji2021show}. We also compare with several online distillation approaches that do not use pre-trained teacher, including DML~\cite{ying2018DML}, CL-ILR~\cite{SongC18}, ONE~\cite{LanZG18}, Snapshot-KD~\cite{YangXSY19}, OKDDip~\cite{chen2020online} and Self-KD~\cite{ZhangSGCBM19}. Then, ablation experiments are conducted to study the impact of evolutionary teacher, guided modules and component of loss function. Finally, we further conduct the experiments on low-resolution and few-sample scenarios to verify the adaptability of our approach. In our experiments, we use VGG(bn)~\cite{SimonyanZ14a}, ResNet\cite{HeZRS16}, Wide ResNet~\cite{ZagoruykoK16}, ShuffleNetV1~\cite{zhang2018shufflenet} and ShuffleNetV2~\cite{ma2018shufflenet} as the backbone networks.

\myPara{CIFAR10.}~The dataset consists of 60,000 $32 \times 32$ colour images in 10 classes, with 6,000 images per class. There are 50,000 training images and 10,000 test images. 

\myPara{CIFAR100.}~This dataset is just like the CIFAR10, except it has 100 classes containing 600 images each. There are 500 training images and 100 testing images per class.

\myPara{Tiny-ImageNet.}~It is an image classification dataset, which contains about $64 \times 64$ sized 100,000 training images and 10,000 verification images with 200 classes. It is more difficult than CIFAR100 dataset. On Tiny-ImageNet, we first randomly adjust the crop size to $224 \times 224$, apply random horizontal flip and normalize it finally . For testing, we only resize the pictures to $224 \times 224$.

\myPara{UMDFaces.}~ This face dataset is collected from Internet, and contains 367,888 face annotations for 8,277 subjects. In our Experiments~\ref{Adaptability Analysis} \textit{Adaptability Analysis}, UMDFaces is as a training dataset. 

\myPara{UCCS.} ~This dataset contains16,149 images in 1,732 subjects in the wild conditions. It is a very challenging benchmark with various levels of challenges, including blurry image, occluded appearance and bad illumination. We follow the setting as~\cite{ge2018low}, randomly select a 180-subject subset, and separate the images into 3,918 training images and 907 testing images, and report the results with the standard top-1 accuracy. 

\begin{figure*}[t]
	\centering
	\subfloat[Baseline]{\includegraphics[width=0.24\linewidth]{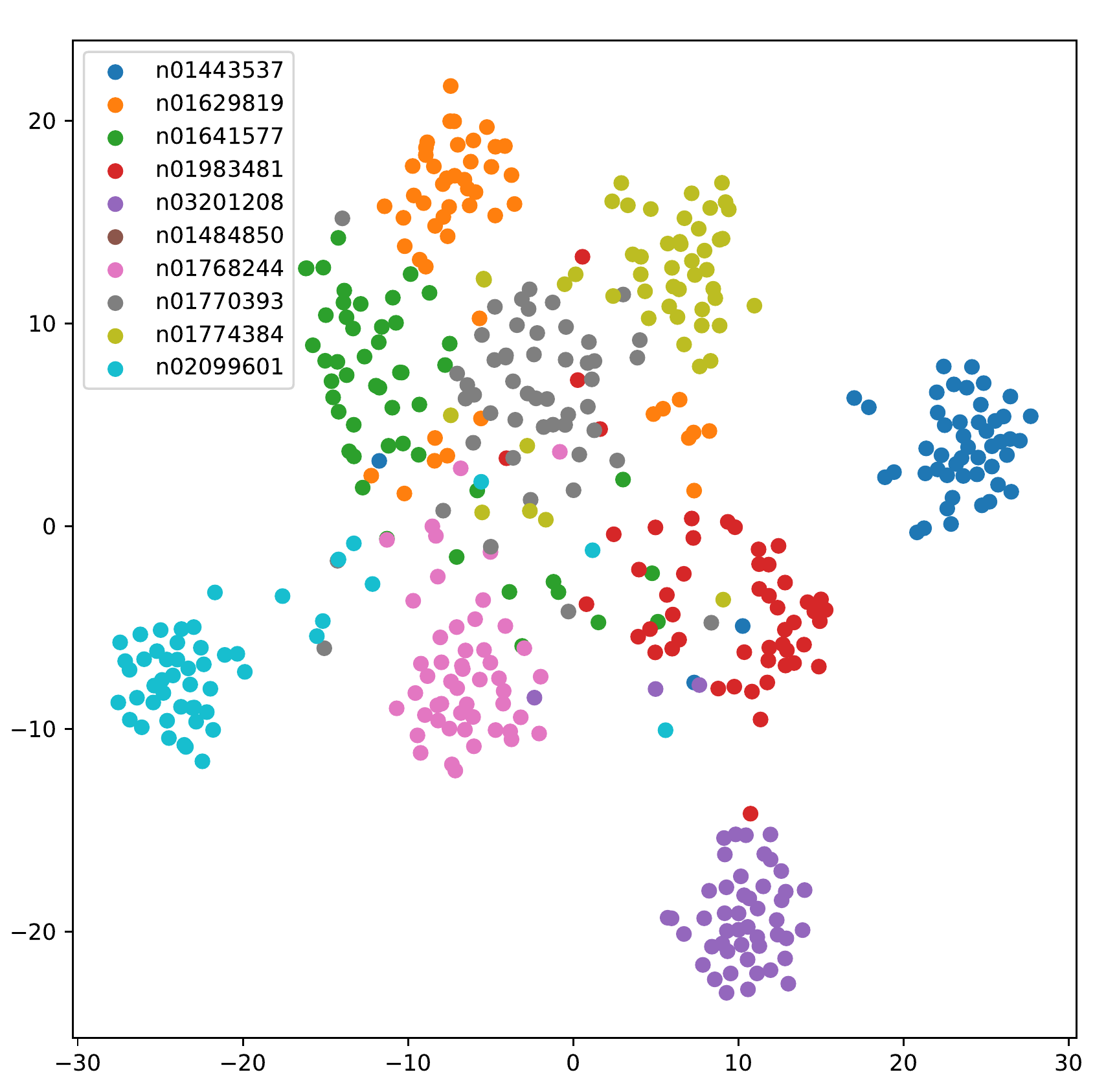}}
	\subfloat[KD]{\includegraphics[width=0.24\linewidth]{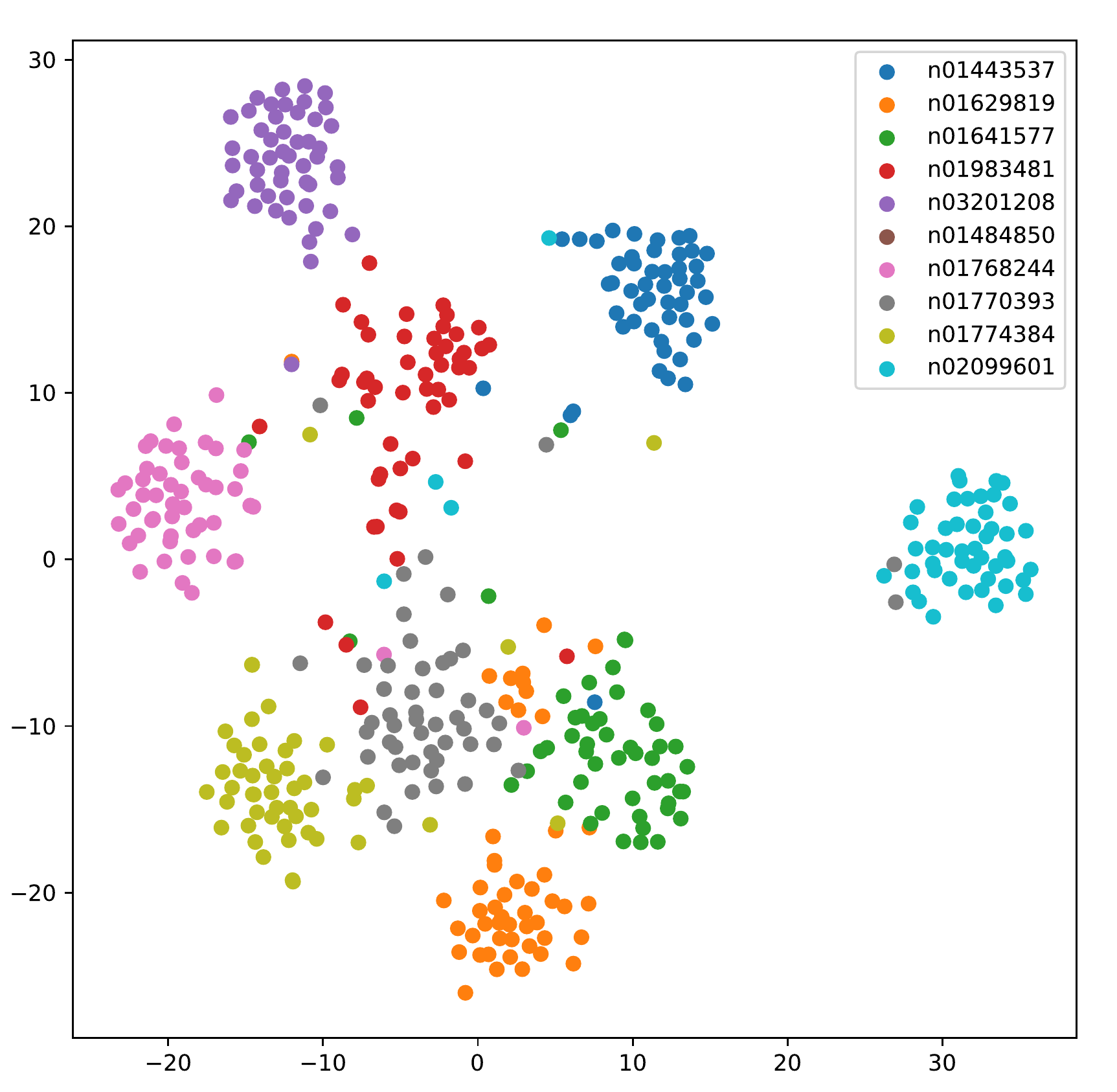}}
	\subfloat[CRD]{\includegraphics[width=0.24\linewidth]{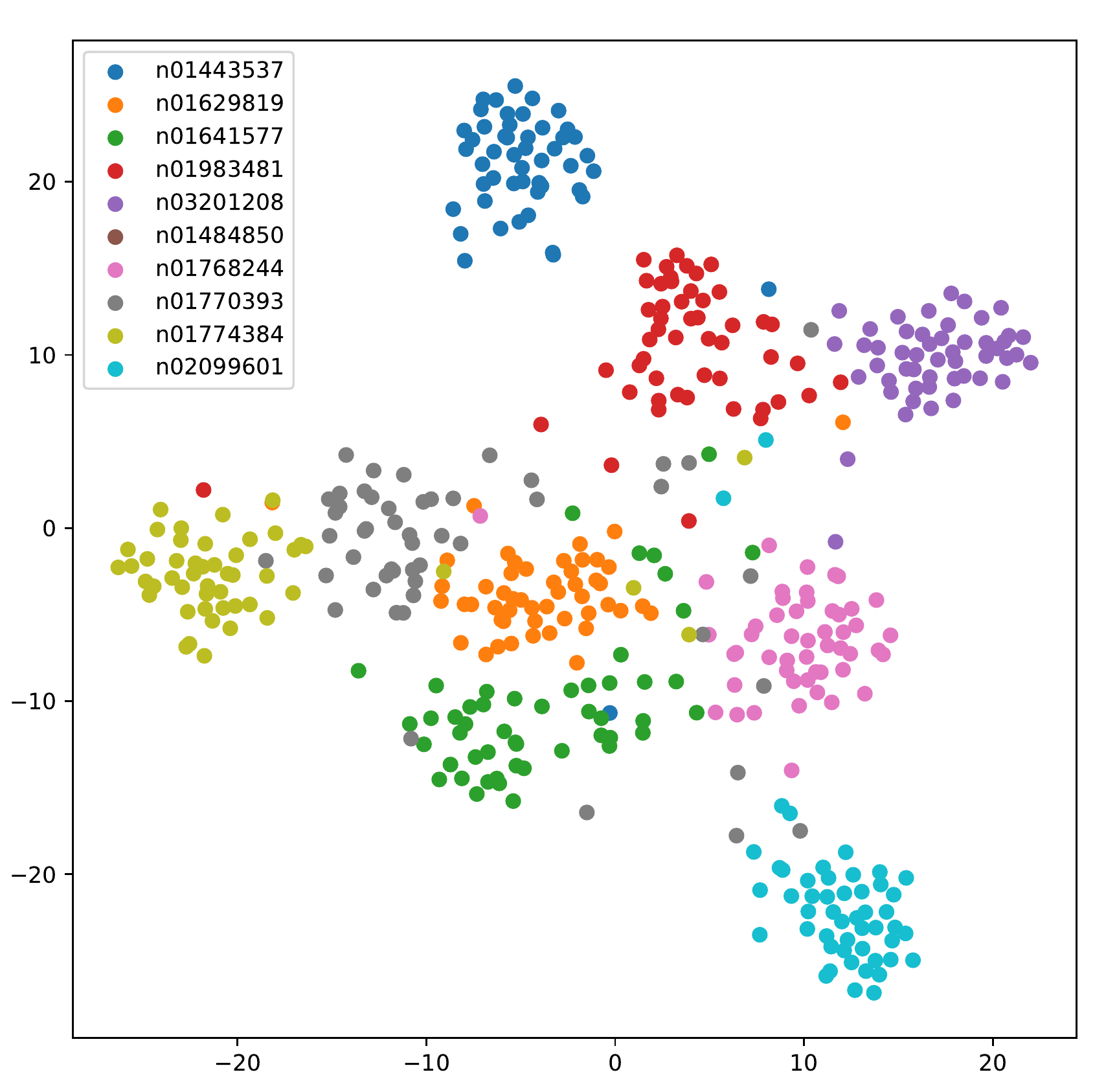}}
	\subfloat[EKD]{\includegraphics[width=0.24\linewidth]{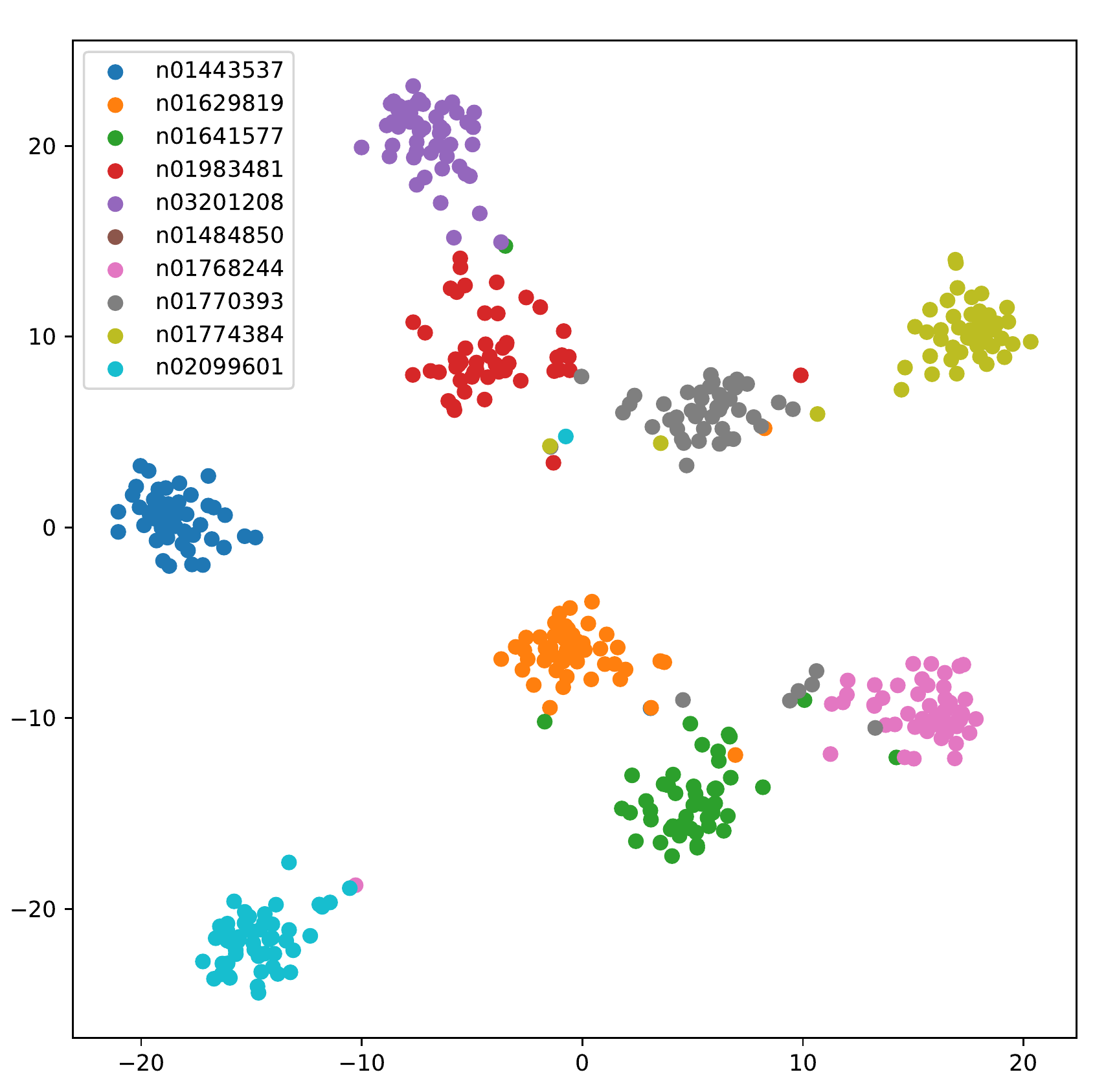}}
	
	\caption{t-SNE Visualisation of ResNet18 trained with KD~\cite{Hinton2015DistillingTK}, CRD~\cite{tian2019crd} and EKD on the Tiny-ImageNet dataset. Different numbers indicate different classes.}
	\label{fig:vis}
\end{figure*}

\begin{table*}[htbp]
	\centering
	\caption{Classification accuracy (\%) on CIFAR100 with different online distillation approaches.}
	\renewcommand{\arraystretch}{1.1}
	\begin{tabular}{c|ccccccccc}
		\hline
		Network & Baseline& DML~\cite{ying2018DML} & CL-ILR~\cite{SongC18} & Snapshot-KD~\cite{YangXSY19} & ONE~\cite{LanZG18} & Self-KD~\cite{ZhangSGCBM19} & OKDDip~\cite{chen2020online}  & EKD (Ours)  \\
		\hline
		VGG16(bn) & 73.81  & 74.67 & 74.38 & - & 74.37 & 75.38 & 75.12 & \textbf{76.86} \\
		ResNet110 &  75.88  & 77.50 & 78.44 &73.51 & 78.33 & 77.04 & 78.91 & \textbf{79.28} \\
		\hline
	\end{tabular}
	\label{online}
\end{table*}

\begin{table*}[htbp]
	\centering
	\caption{Performance comparison on Tiny-ImageNet. The teacher is ResNet34 and the student is ResNet18.}
	\renewcommand{\arraystretch}{1.1}
	\begin{tabular}{c|cccccccccc}
		\hline
		Method & Student & KD~\cite{Hinton2015DistillingTK} & FitNet~\cite{RomeroBKCGB14} & Attention~\cite{ZagoruykoK17}  & RKD~\cite{park2019relational} & CRD~\cite{tian2019crd} &  Res-KD~\cite{li2020reskd} & AFD~\cite{ji2021show} & EKD (Ours)  \\
		\hline
		Accuracy (\%)  & 65.30 & 68.18 & 67.79 & 67.82 & 67.72 & 68.19 & 68.62 & 68.80 &\textbf{69.46} \\
		\hline
	\end{tabular}
	\label{tinyimagenet}
\end{table*}


\subsection{Comparisons with Offline Distillation Approaches}
We conduct experiments on CIFAR100 dataset and perform the comparisons on the offline distillation with peer-architecture setting and cross-architecture setting. The experimental results of other offline distillation methods are based on the open source code \textit{RepDistiller} of CRD~\cite{tian2019crd}, we implement experiments according to their hyper-parameters setting.

\myPara{Peer-Architecture Setting.}~In this setting, the teacher and student networks share similar or the same structures. From the results shown in Tab.~\ref{tab:similararch}, we get some meaningful observations. First, our approach outperforms all other offline distillation approaches, implying its remarkable effectiveness in improving student network learning, such as, when the teacher and student is VGG19(bn) and VGG11(bn) respectively, we achieve 74.12\% accuracy on CIFAR100 which is 0.63\% higher than the state-of-art method Similarity~\cite{TungM19}, and we gain 3.13\% improvement when the teacher and student is ResNet50 and ResNet18 compared to CRD~\cite{tian2019crd}. Second, the effect is also obvious when the teacher stream and student stream adopt the same structure. For example, when both the teacher and student are VGG11(bn), we gain 1.20\% improvement even higher than learning from VGG19(bn), and the same is true when teacher and student are ResNet50. We speculate that it's due to the smaller capability gap between two identical backbone networks, which facilitates the knowledge transfer from evolutionary teacher to student.

\myPara{Cross-Architecture Setting.}~According to the experimental results of the previous section, we can find that the effect of proposed approach in the same or similar network architecture is superior to the conventional offline knowledge distillation methods. Then, to verify the generalization performance of our approach, we conduct further experiments on more challenging knowledge transfer tasks, cross-architecture setting, which means the architecture of teacher and student network are completely different. 

We verify the performance of cross-architecture setting on CIFAR100 dataset, and learning rate, update strategy and simple data augmentation methods keep the same settings as before. The results are shown in Tab.~\ref{tab:crossarch}. We achieve the best results under the condition that the teacher and student networks are completely different. Specifically, when the teacher is ResNet18 and the student is VGG8(bn), our approach is at least 0.65\% higher than the other offline distillation strategies, and for the teacher is WRN50-2, the student (VGG8 (bn)) gains a clear advantage, its performance improvement is more than 1.55\%. When the teacher and student network are WRN50-2 and ShuffleNetV1 respectively, we achieve an improvement of at least 1.83\% except when compared to Similarity~\cite{TungM19}. These are due to the evolutionary teacher reduces the shackles caused by the capability gap between teacher and student, provides richer supervision information, and the guided modules ensure the efficiency of knowledge representation and transfer. Results on cross-architecture setting clearly demonstrate that our method does not rely on architecture-specific cues.

\myPara{Comparisons on Tiny-ImageNet.}
In order to further verify the effectiveness of our proposed method, we conduct experiments on the more challenging Tiny-ImageNet. 
We mainly compare our approach with some typical methods, including KD~\cite{Hinton2015DistillingTK}, FitNet~\cite{RomeroBKCGB14}, Attention~\cite{ZagoruykoK17}, RKD~\cite{park2019relational}, CRD~\cite{tian2019crd}, Res-KD~\cite{li2020reskd}, AFD~\cite{ji2021show}.
Here, Res-KD used the knowledge gap between teacher and student as a guide to train a more lightweight student network, which we call ``res-student", and AFD is an attention-based feature matching distillation method utilizing all the feature levels of the teacher. As illustrated with Tab. \ref{tinyimagenet}, the proposed method EKD achieves better performance on large-scale datasets than other baseline knowledge distillation methods, for example, our method achieves a 0.66\% performance improvement compared with the previous best results of AFD.

\subsection{Comparisons with Online Distillation Approaches}
We also compare the proposed EKD to several recent online knowledge distillation approaches, including Deep Mutual Learning (DML)~\cite{ying2018DML}, Collaborative Learning for Deep Neural Networks (CL-ILR)~\cite{SongC18}, Knowledge Distillation by On-the-Fly Native Ensemble (ONE)~\cite{LanZG18}, Snapshot-KD~\cite{YangXSY19}, Self Distillation (Self-KD)~\cite{ZhangSGCBM19} and Online knowledge distillation with diverse peers (OKDDip)~\cite{chen2020online}. The results in Tab.~\ref{online} are based on the code of OKDDip~\cite{chen2020online}, in which the results of Snapshot-KD are based data from~\cite{YangXSY19}, and the results of Self-KD are obtained by re-implementing the framework of original research paper~\cite{ZhangSGCBM19}.

As shown in the Tab.~\ref{online}, the “Baseline” approach trains a model by ground-truth labels only. Compared with the other online knowledge distillation methods, our EKD shows some advantages, when the backbone network is VGG16(bn), our method is 1.48\% higher in accuracy than the current best approach Self-KD~\cite{ZhangSGCBM19}. And when the backbone network is ResNet110, our method has a 0.37\% improvement than OKDDip~\cite{chen2020online}, which is an increase of 3.40\% from the “Baseline”. We suspect that the capability gap still exits due to the lack
of representation in detail and the process of learning for previous online knowledge distillation methods. So that the insufficient and relatively unreliable supervision information from peers or itself will restrict the learning of student network to some extent. Our evolutionary teacher not only reduces the capability gap brought by fixed pre-trained teacher, but also provides a more flexible teacher role for online knowledge distillation instead of themselves or their peers. Above results illustrate that our method can transfer knowledge more adequately and effectively by combining guided modules and evolutionary teacher.

\subsection{Visualisation}
Previous experimental results quantitatively demonstrate the superiority of our proposed evolutionary knowledge distillation. In order to further visually demonstrate the advantages of our approach, we use the t-SNE~\cite{van2008visualizing} for visualization. t-SNE is a tool to visualize high-dimensional data. It converts similarities between data points to joint probabilities and tries to minimize the Kullback-Leibler divergence between the joint probabilities of the low-dimensional embedding and the high-dimensional data. The Fig.~\ref{fig:vis} gives the visualisation results of ResNet18 trained with KD~\cite{Hinton2015DistillingTK}, CRD~\cite{tian2019crd} and EKD on the Tiny-ImageNet dataset. The “Baseline” in Fig.~\ref{fig:vis} denotes that we train ResNet18 without any distillation methods, for KD, CRD and EKD, we train the student with the helps of teacher ResNet50. We randomly select ten classes in this dataset for visualization experiment, with 50 samples for each class, different numbers indicate different classes in Fig.~\ref{fig:vis}. To begin with, it is obvious that our approach achieves more concentrated clusters than KD and CRD. In addition, as demonstrated in Fig.~\ref{fig:vis}, the changes of the distances in classifiers of KD and CRD are more severe than that in classifier of EKD. We speculate that the evolutionary teacher can facilitate the student network learning by narrowing the teacher-student capability gap and the guided modules can enhance intermediate knowledge representation to improvement of student network learning.

\subsection{Further Analysis}
The approach we proposed mainly includes two parts, evolutionary teacher and guided modules. In this section, we conduct further experiments to explore the specific influence of evolutionary teacher, guided modules and components of loss function. In addition, we compare the influence of our approach in training time with other distillation approaches. All ablation experiments were performed in the same setting as previous ones.

\begin{table}[htbp]
	\centering
	\caption{Classification accuracy on CIFAR100. The teacher is ResNet50 and the student is ResNet18. ``ET'' denotes evolutionary teacher without guided modules; ``T\_G'' denotes teacher stream with guided modules; ``S\_G'' denotes student stream with guided modules.}
	\renewcommand{\arraystretch}{1.1}
	\begin{tabular}{c|c}
		\hline
		Approach &Accuracy (\%) \\
		\hline
		KD &  77.57\\
		KD+ET&  79.71 (+2.14) \\
		\hline
		KD+S\_G & 78.85 \\
		KD+S\_G+ET & 81.34 (+2.49) \\
		\hline
		KD+T\_G+S\_G  & 80.01 \\
		KD+T\_G+S\_G+ET (EKD) & 82.05 (+2.04)\\
		\hline
	\end{tabular}
	\label{ablation}
\end{table}

\begin{figure}[h]
	\centering
	\includegraphics[width=1.0\linewidth]{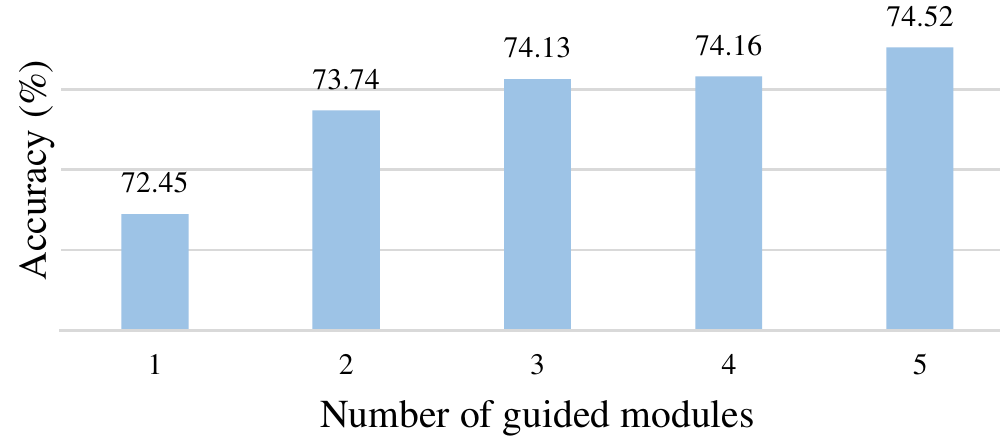}
	\caption{The influence of guided modules. We train student models with different numbers of guided module pairs on CIFAR100 dataset.}
	\label{fig:blocks}
\end{figure}

\myPara{Influence of Evolutionary Teacher.} Here, we make ResNet50 and ResNet18 as teacher and student, respectively. As illustrated in Tab. \ref{ablation}, for the classic knowledge distillation method KD~\cite{Hinton2015DistillingTK}, the accuracy of student on the CIFAR100 dataset is 77.57\%. When we use an evolutionary teacher to teach student network, and other settings remain unchanged, the performance of our student network will increase by 2.14\%, reach 79.71\%. We believe it is because the evolutionary teacher provides more sufficient supervision information to promote student network learning. What's more, the guided modules play a positive role in within-stream distillation and cross-stream distillation, so, when the guided modules are working, can the evolutionary teacher still play an active role in the learning of student? In order to further verify the influence of evolutionary teacher, we added guided modules to teacher stream and student stream respectively, and compared the performance differences of student model before and after adopting evolutionary teacher strategy. As shown in Tab. \ref{ablation}, Introducing the guided modules for the student stream only, the evolutionary teacher strategy improves the accuracy by 2.04\%, while the introduction of the guided modules for the teacher and student streams, the evolutionary teacher strategy improves the performance by 2.49\%. It's necessary to point out that the teacher model also has a good performance with the help of its guided modules, when our evolutionary teacher without guided modules to supervise the learning of student, it can still increase by 3.77\% compared to KD (81.37\% VS. 77.57\%). The above results demonstrate that evolutionary teacher can promote student network learning by reducing the capability gap between teacher and student.

\myPara{Influence of Guided Modules.} In order to explore the influence of introduced guided modules, we conduct further experiments. As shown in Tab. \ref{ablation}, The effect of the guided modules to improve the performance of the student network is obvious. When only the student stream uses the guided modules, the performance of student is improved by 1.28\% compared to KD (78.85\% VS. 77.57\%). When the teacher stream and student stream use the guided modules at the same time, the performance is improved by 2.44\% (80.01\% VS. 77.57\%). These results show that the guided modules promote the effective transfer of knowledge by making full use of the intermediate information of the network, and improve the performance of student. 

In addition, the number of guided modules also has an important impact on knowledge transfer, so, we conduct experiments base on VGG11(bn). For the network structure, we use up to four guided modules and at least zero module. When the number of guided modules is zero, our approach essentially degenerates into classic knowledge distillation with evolutionary teacher. As shown in Fig. \ref{fig:blocks}, experimental results show that guided modules play a vital role. When the number of guided modules increases, the effect is continuously improved, but as the number increases, the improvement becomes less obvious. It should be pointed out that due to the extremely simple structure of guided modules, their consumption of computing resources is negligible.

\begin{figure}[h]
	\centering
	\includegraphics[width=1.0\linewidth]{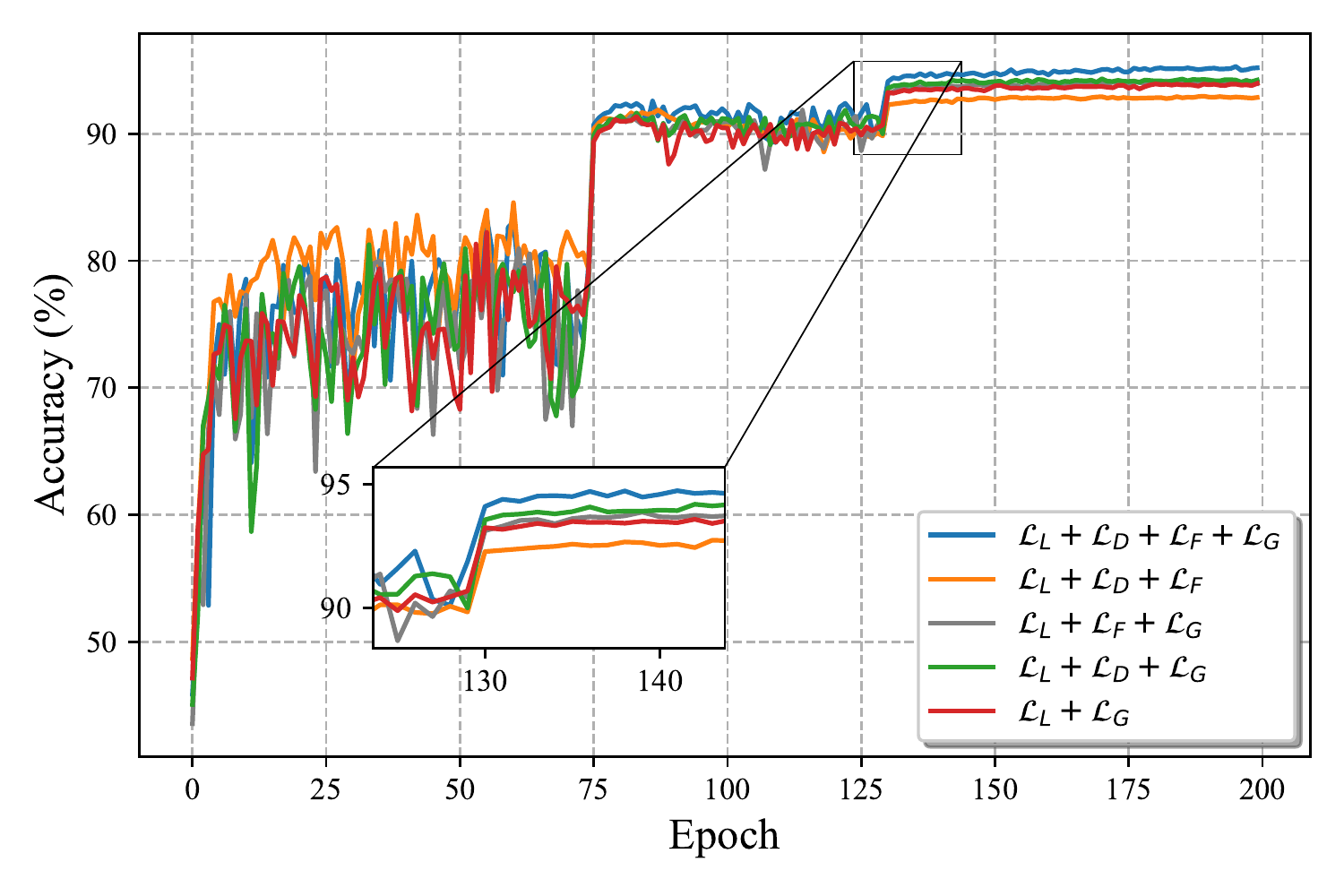}
	\caption{Ablation study about components of loss function on CIFAR10 dataset. The teacher and student are VGG19(bn) and VGG11(bn) respectively.}
	\label{fig:ablation}
\end{figure}

\myPara{Influence of Loss Function.} The training process of EKD includes two simultaneous stages, within-stream distillation and cross-stream distillation. The loss function mainly includes within-stream distillation loss ($\mathcal L_{D}$, $\mathcal L_{F}$) and cross-distillation loss ($\mathcal L_{G}$), $\mathcal L_{L}$ is classification loss of each classifier. We conduct related experiments to explore the influence of components of loss function (Eq.~(\ref{total})). The results are shown in the Fig.~\ref{fig:ablation}, it is obvious that the loss function $\mathcal L_{G}$ has a promotion effect on the improvement of student network, which shows that the guided modules effectively promote the evolutionary teacher to transfer knowledge to student. $\mathcal L_{D}$ and $\mathcal L_{F}$ are the distillation losses that act on the within-stream feature level and predicted distribution level, respectively. All of them play a positive role in the learning of the student network, and the effect of $\mathcal L_{D}$ is more obvious, which indicate that the softened label distribution and feature maps that contain rich information about image intensity and spatial correlation provide sufficient supervision for the learning of students in within-stream and cross-stream distillation. In general, for within-stream distillation, deeper classifiers provide supervision to help the learning of shallow ones, thereby bringing about the performance improvement of the stream itself. Cross-stream distillation can improve knowledge transfer from the evolutionary teacher with the help of guided modules. 

\begin{figure}[h]
	\centering
	\includegraphics[width=1.0\linewidth]{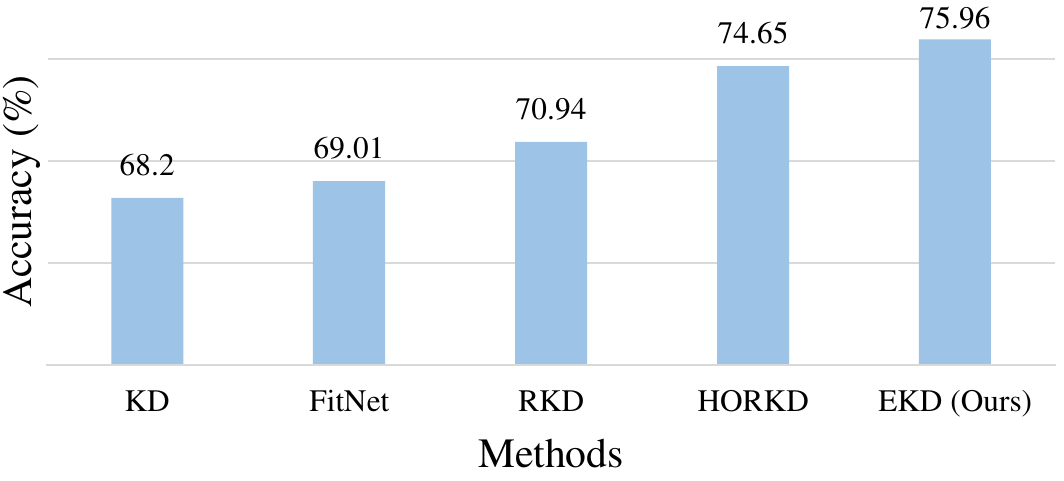}
	\caption{Low-resolution ($16\times16$) image classification accuracy on CIFAR100 dataset.}
	\label{LR}
\end{figure}

\myPara{Training Time Analysis.} We compare training time with other knowledge distillation methods, all the comparative experiments are implemented on NVIDIA TITAN GPU with identical hyper-parameters. In fact, the guided modules won't increase the amount of calculation too much because of their extremely simple structure, what's more, since the traditional distillation method needs to train the teacher and student model separately in two stages, our approach trains them at the same time, so the training time will not increase significantly. Specifically, when the teacher and student are ResNet50 and ResNet18 respectively, the training time of our approach is 277.5s/epoch, the traditional knowledge distillation method KD~\cite{Hinton2015DistillingTK} training time for teacher and student are 205.29s/epoch and 171.07s/epoch. The memory of GPU occupied during training is slightly higher than that of traditional distillation methods. In general, our training time is not significantly higher than traditional methods, but brings considerable performance improvements.

\subsection{Adaptability Analysis}\label{Adaptability Analysis}
\myPara{Low-resolution Scenario.}~To verify the performance of EKD in low-resolution scenario, we conduct experiments on challenging low-resolution image classification and low-resolution face recognition tasks. 

For image classification task, we first train ResNet50 as teacher, and then make ResNet18 as student. The dataset used is down-sampled to reduce the resolution to 16$\times$16, and the classification loss is classic cross-entropy loss.The experimental results are shown in Fig.~\ref{LR}, our approach shows good adaptability on more challenging low-resolution image classification task, which is 1.31\% higher than the current best method HORKD~\cite{GeHORKD20}. In addition, HORKD needs to consume more computing resources due to the introduction of the assistant network. These results indicate the effectiveness of evolutionary teacher supervision and the introduced guided modules in representing and transferring knowledge in low-resolution scenario.

\begin{table}[htbp]
	\centering
	\caption{The performance of various methods on UCCS benchmark. Our student model achieves great performance when working at low resolution ($ 16\times16 $) and costing less parameters.}
	\renewcommand{\arraystretch}{1.1}
	\begin{tabular}{l|c|c|c}
		\hline
		Method & Top-1 Accuracy (\%) & Parameters & Year \\
		\hline
		VLRR~\cite{7780887}     & 59.03  &  {-} & 2016 \\
		SphereFace~\cite{liu2017sphereface} & 78.73 & 37M & 2017 \\
		CosFace~\cite{wang2018cosface} & 91.83 & 37M & 2018 \\
		SKD~\cite{ge2018low}   & 67.25 &  {0.79M} & 2019 \\
		AGC-GAN~\cite{talreja2019attribute}   & 70.68 &  -  & 2019 \\
		VGGFace2~\cite{CaoSXPZ18}   & 84.56 &  26M  & 2019 \\
		ArcFace~\cite{deng2019arcface}  & 88.73 &  37.8M  & 2019 \\
		LRFRW~\cite{li2019low}   & 93.40 & 4.2M  & 2019 \\
		HORKD~\cite{GeHORKD20}   & 92.11 & {7.8M} & 2020 \\
		\textbf{EKD (Ours)} & \textbf{93.85}  & \textbf{0.61M} & - \\
		\hline
	\end{tabular}
	\label{tab:uccs}
\end{table} 
\begin{figure}[h]
	\centering
	\includegraphics[width=1.0\linewidth]{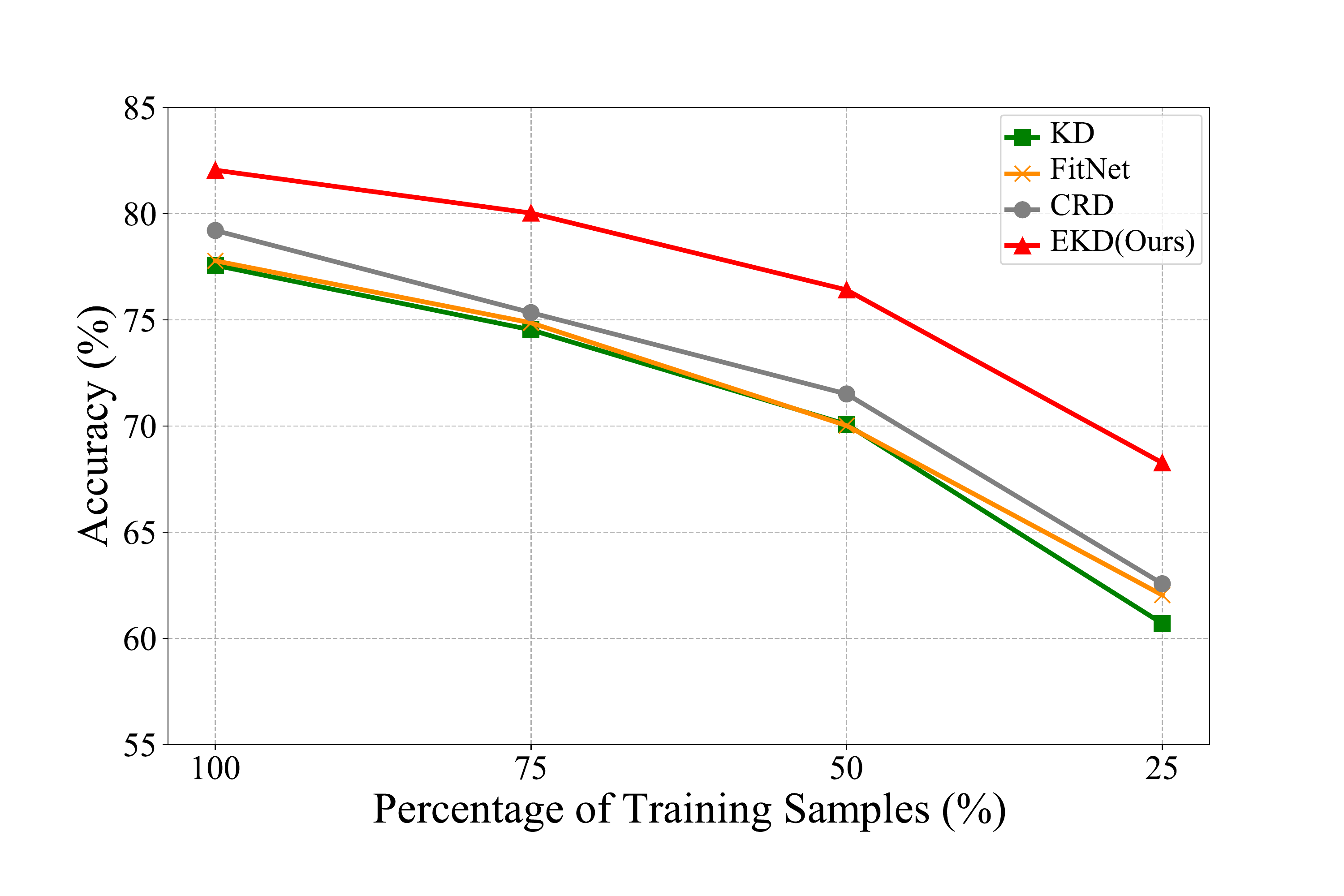}
	\caption{Few-sample image classification results on CIFAR100 dataset. EKD trains teacher and student with subsets. KD, FitNet and CRD train the teacher models on full set and train the students with subsets.}
	\label{fig:Few-shot}
\end{figure}

Then we conduct experiments on the low-resolution face recognition task which is very helpful in many real-world applications, \eg, recognizing low-resolution surveillance faces in the wild. In our experiments, the teacher uses a recent face recognizer VGGFace2~\cite{CaoSXPZ18} with ResNet50 structure and the student network is based on streamlined ResNet18 with only 0.61M parameters, they are trained on UMDFaces~\cite{bansal2017umdfaces} and tested on UCCS dataset~\cite{uccs}. In order to verify the validity of our low-resolution models, we emphatically check the accuracy when the input resolution is 16$\times$16.
As shown in Tab.~\ref{tab:uccs}, our student model achieves better low-resolution face recognition performance and costs less parameters. Specifically, we achieve a top-1 accuracy of 93.85\% on the UCCS benchmark, which is 0.45\% higher than the state-of-art LRFRW~\cite{li2019low}, and the amount of parameters is reduced by nearly ten times. Classical face recognition methods, such as ArcFace and CosFace, do not perform well in low-resolution scenario, with their highest recognition accuracy only reaching 88.73\%. Moreover, their models have more parameters, which will lead to a significant increase in the computing cost for inference. It is worth mentioning that whether it is VLRR~\cite{7780887}, SKD~\cite{ge2018low}, HORKD~\cite{GeHORKD20} or LRFRW~\cite{li2019low} in the distillation process, high-resolution images corresponding to low-resolution faces are necessary to provide more information, but such high-resolution images are not always easy to obtain, our approach only uses low-resolution face images for training, which adopts real-world application scenarios.

\myPara{Few-sample Scenario.}~In practical scenario (\eg, in the wild), the number of samples available for training is usually limited to a certain extent. In order to study the adaptability of our approach in a few-sample scenario, we conduct experiments on different subsets of CIFAR100 dataset. We randomly select images of each class to form new subsets, and use the newly designed training set to train the student models while maintaining the same test set. ResNet50 and ResNet18 are chosen as teacher and student, respectively. We compare the performance of student models with several typical distillation methods include KD~\cite{Hinton2015DistillingTK}, FitNet~\cite{RomeroBKCGB14} and CRD~\cite{GeHORKD20}. The percentages of retained samples are 100\%, 75\%, 50\% and 25\%. For a fair comparison, we use the same data in different distillation approaches to train student models and train their teacher models on full dataset. 

The results in Fig.~\ref{fig:Few-shot} show that our approach remarkably surpasses other distillation approaches under few-sample scenario, even when the teacher stream of EKD is learned and distilled knowledge from less training data. Specifically, as the amount of training data decreases, the performance of distillation methods represented by KD, FitNet, and CRD in the figure will decrease significantly, while the downward trend of EKD proposed by us is significantly milder, even the advantage is more obvious when the training samples are less. When the percentage of training samples is 25\%, our approach is nearly 10\% higher than CRD.

\section{Conclusion}\label{sec5}
In this paper, we propose an evolutionary knowledge distillation and show its superiority by comparing it with the state-of-the-art offline distillation and online distillation approaches. Our approach uses an evolutionary teacher to supervise the learning of student from scratch, which can reduce the teacher-student capability gap to promote knowledge transfer. What's more, through the introduction of some simple guided module pairs between corresponding teacher-student blocks, the efficiency of intermediate knowledge representation is improved. We believe this evolutionary knowledge transfer manner and simple guided mechanism are very promising in knowledge distillation community. In the future, we will explore the potential of our approach in combining with existing knowledge distillation schemes, and performing more practical tasks.

\myPara{Acknowledgement}. This work was partially supported by grants from the National Key Research and Development Plan (2020AAA0140001), National Natural Science Foundation of China (61772513), Beijing Natural Science Foundation (19L2040), and the project from Beijing Municipal Science and Technology Commission (Z191100007119002). Shiming Ge is also supported by the Youth Innovation Promotion Association, Chinese Academy of Sciences.

\bibliographystyle{IEEEtran}
\bibliography{bibieee}

%
%
%
%
%
%
%

%

\vspace{-1cm}
\begin{IEEEbiography}[{\includegraphics[width=1in,height=1.25in,clip,keepaspectratio]{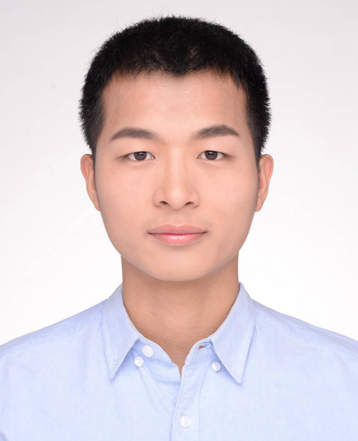}}]{Kangkai Zhang} received his B.S. degree in Electronical Information Science and Technology from the School of Electronic Information and Optical Engineering in Nankai University, Tianjin, China. He is now a Master student at the Institute of Information Engineering at Chinese Academy of Sciences and the School of Cyber Security at the University of Chinese Academy of Sciences, Beijing. His major research interests are deep learning and computer vision.
\end{IEEEbiography}
\vspace{-1cm}
\begin{IEEEbiography}[{\includegraphics[width=1in,height=1.25in,clip,keepaspectratio]{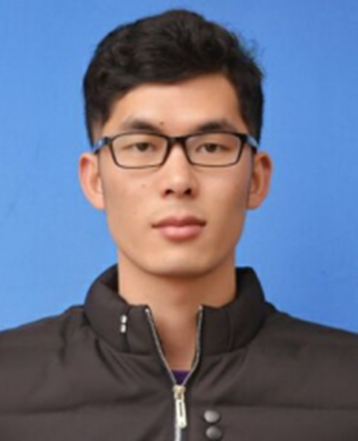}}]{Chunhui Zhang} received the B.S. degree from the School of Mathematics and Computational Science, Hunan University of Science and Technology, Xiangtan, China. He is currently pursuing the master’s degree with the Institute of Information Engineering, Chinese Academy of Sciences, Beijing, China, and the School of Cyber Security, University of Chinese Academy of Sciences, Beijing. His major research interests include machine learning and visual tracking.
\end{IEEEbiography}
\vspace{-1cm}
\begin{IEEEbiography}[{\includegraphics[width=1in,height=1.25in,clip,keepaspectratio]{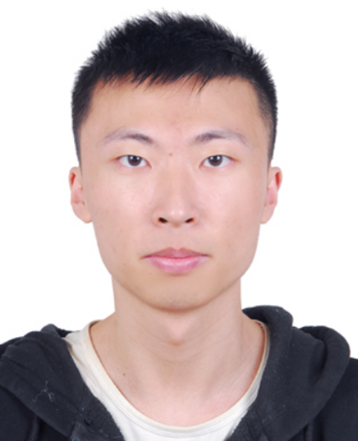}}]{Shikun Li} received the B.S. degree from the School of Information and Communication Engineering, Beijing University of Posts and Telecommunications (BUPT), Beijing, China. He is currently pursuing the Ph.D. degree with the Institute of Information Engineering, Chinese Academy of Sciences, Beijing, and the School of Cyber Security, University of Chinese Academy of Sciences, Beijing. His research interests include machine learning, data analysis, and computer vision.
\end{IEEEbiography}
\vspace{-1cm}
\begin{IEEEbiography}[{\includegraphics[width=1in,height=1.25in,clip,keepaspectratio]{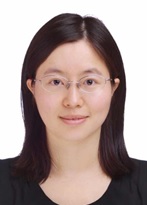}}]{Dan Zeng} received her Ph.D. degree in circuits and systems, and her B.S. degree in electronic science and technology, both from University of Science and Technology of China, Hefei. She is a full professor and the Dean of the Department of Communication Engineering at Shanghai University, directing the Computer Vision and Pattern Recognition Lab. Her main research interests include computer vision, multimedia analysis, and machine learning. She is serving as the Associate Editor of the IEEE Transactions on Multimedia, the TC Member of IEEE MSA and Associate TC member of IEEE MMSP.
\end{IEEEbiography}
\vspace{-0.8cm}
\begin{IEEEbiography}[{\includegraphics[width=1in,height=1.25in,clip,keepaspectratio]{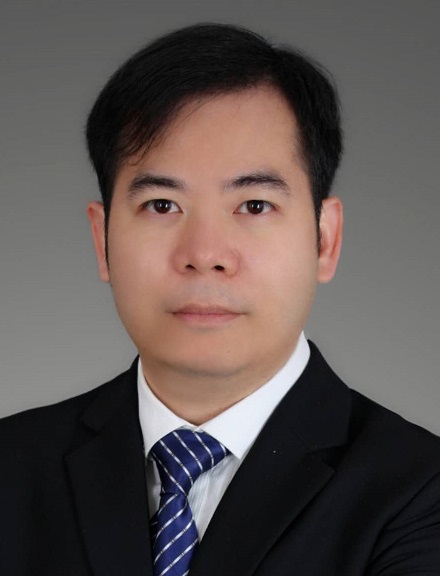}}]{Shiming Ge} (M'13-SM'15) is a Professor with the Institute of Information Engineering, Chinese Academy of Sciences. He is also the member of Youth Innovation Promotion Association, Chinese Academy of Sciences. Prior to that, he was a senior researcher and project manager in Shanda Innovations, a researcher in Samsung Electronics and Nokia Research Center. He received the B.S. and Ph.D degrees both in Electronic Engineering from the University of Science and Technology of China (USTC) in 2003 and 2008, respectively. His research mainly focuses on computer vision, data analysis, machine learning and AI security, especially efficient learning solutions towards scalable applications. He is a senior member of IEEE, CSIG and CCF.
\end{IEEEbiography}




\end{document}